\documentclass[conference]{IEEEtran}
\IEEEoverridecommandlockouts

\usepackage{cite}
\usepackage{amsmath,amssymb,amsfonts}
\usepackage{hyperref}
\usepackage{algorithmic}
\usepackage{graphicx}
\usepackage{textcomp}
\usepackage{xcolor}
\usepackage{booktabs, multirow}
\def\BibTeX{{\rm B\kern-.05em{\sc i\kern-.025em b}\kern-.08em
    T\kern-.1667em\lower.7ex\hbox{E}\kern-.125emX}}
\begin{document}


\title{Modeling Uncertainty: Constraint-Based Belief States in Imperfect-Information Games\thanks{Short Paper}}


\author{
    \IEEEauthorblockN{Achille Morenville}
    \IEEEauthorblockA{\textit{ICTEAM, UCLouvain} \\
    Louvain-la-Neuve, Belgium \\
    achille.morenville@uclouvain.be}
    \and
    \IEEEauthorblockN{Éric Piette}
    \IEEEauthorblockA{\textit{ICTEAM, UCLouvain} \\
    Louvain-la-Neuve, Belgium \\
    eric.piette@uclouvain.be}
}

\maketitle

\begin{abstract}

    In imperfect-information games, agents must make decisions based on partial knowledge of the game state. The Belief Stochastic Game model addresses this challenge by delegating state estimation to the game model itself. This allows agents to operate on externally provided belief states, thereby reducing the need for game-specific inference logic. This paper investigates two approaches to represent beliefs in games with hidden piece identities: a constraint-based model using Constraint Satisfaction Problems and a probabilistic extension using Belief Propagation to estimate marginal probabilities. We evaluated the impact of both representations using general-purpose agents across two different games. Our findings indicate that constraint-based beliefs yield results comparable to those of probabilistic inference, with minimal differences in agent performance. This suggests that constraint-based belief states alone may suffice for effective decision-making in many settings.
    
\end{abstract}

\begin{IEEEkeywords}
General Game Playing, Imperfect-Information Games, Knowledge Representation.
\end{IEEEkeywords}

\section{Introduction}

    Imperfect-information games pose a key challenge in Artificial Intelligence (AI), requiring agents to make decisions under uncertainty about the true state of the game. Unlike perfect-information games, where all aspects of the environment are observable, these games involve hidden elements that players must infer, such as unknown cards or secret piece positions. The inference process is complex and usually requires agents to maintain internal estimates of the actual state, often relying on handcrafted game-specific logic. These solutions hinder the development of general and reusable agents, particularly in the context of General Game Playing (GGP)~\cite{Genesereth_2005_GGP}.

    To address this challenge, the Belief Stochastic Game (Belief-SG) model was recently introduced~\cite{Morenville_2024_BeliefSG}. This model shifts the responsibility for state estimation from the agent to the game model itself, enabling the development of domain-independent agents. In Belief-SG, the agent receives a belief state from the model, representing a probability distribution over the actual state of the game. This externalization frees agents from having to implement their own inference procedures, allowing them to focus purely on strategy.


    Although Belief-SG allows for detailed modeling of uncertainty, it remains unclear whether such precise estimation is necessary for effective decision-making. In many games, the uncertainty over the state of the game can be tightly constrained using purely logical reasoning without computing explicit probability distributions. This raises the question of whether the complexity of probabilistic inference offers strategic advantages over simpler logic-based models tracking possible states.

    To investigate this question, we leverage a constraint satisfaction problem (CSP)~\cite{Rossi_2006_handbook} formulation of the belief state, which efficiently encodes the set of game states that remain logically possible given past observations. On top of this CSP representation, we implement a Belief Propagation (BP)~\cite{Pearl_2022_Reverend} layer to estimate the likelihoods over the CSP-encoded variables, enabling the computation of probability distributions over hidden information. This structure allows us to compare agents that rely solely on feasibility constraints with those that additionally exploit probabilistic beliefs derived from the same underlying model. Our experimental evaluation examines how these different forms of belief access affect agent performance in the Belief-SG model.

    The aim of this study is to evaluate the practical impact of belief precision on strategic effectiveness. By isolating the informational content available to agents, we aim to better understand the trade-offs involved in belief modeling. These insights contribute to the ongoing development of general agents capable of playing a broad class of imperfect-information games without domain-specific knowledge.

\section{Background}

    Imperfect-information games are commonly formalized using models such as Extensive Form Games (EFG)~\cite{Neumann_1944_Theory} and Factored-Observation Stochastic Games (FOSG)~\cite{Kovavrik_2022_FOSG}. While both models have played a key role in the development of AI agents for imperfect-information games, they have limitations that hinder their application in the context of GGP. 

    In EFG, there is no distinction between public and private information, nor is there a representation of what each player knows about the game state. However, this distinction is crucial for decision-making and search in imperfect-information settings. Although extensions have been proposed to address this issue~\cite{Burch_2014_Solving}, it has been shown that this distinction cannot be extracted in a general way from EFG representations~\cite{Kovavrik_2019_ProblemsEFG}, making the model unsuitable for GGP.

    FOSG was introduced to overcome some of these issues by modeling partial observability explicitly. In this framework, agents do not observe the full game state but receive structured observations, divided into public and private parts, which allows some reasoning about the knowledge of other agents. While FOSG supports sound search under uncertainty~\cite{Sustr_2021_Sound}, agents receive only partial observations about the state of the game, requiring them to construct and maintain state estimates. This often results in handcrafted state estimation tailored to specific games, making it difficult to generalize developments across a wide variety of games.
    
    These limitations have led to highly specialized agents. For example, Libratus~\cite{Brown_2017_Superhuman} and DeepStack~\cite{Moravcik_2017_DeepStack} are tailored for Heads-Up No-Limit Texas Hold’em Poker, with custom abstractions and state representations. Even agents that aim for generality, such as Student of Games~\cite{Schmid_2023_Student}, rely on the selection of game-specific architectures, limiting their true generality.


    The Belief Stochastic Game (Belief-SG) model was recently proposed to address these challenges by externalizing the state estimation process to the game model itself. In Belief-SG, the agent receives from the model a belief state, a probability distribution over the actual state of the game. Public and private information can still be separated by providing different belief states to different agents. This design enables agents to focus purely on strategic decision-making without having to implement game-specific inference procedures. As a result, Belief-SG provides a more general framework for reasoning in imperfect-information games within the context of GGP.

    Due to the limitations of the EFG and FOSG models, most prior work on state estimation in imperfect-information games has relied on handcrafted, game-specific logical inference or probabilistic models. While variants of CSPs have been used for the modeling and the detection of symmetries in perfect-information games~\cite{Koriche_2017_Symmetry}, to our knowledge, constraint-based belief state modeling, and in particular the use of Belief Propagation over CSPs, has not been explored in the context of GGP for imperfect-information games.

\section{Belief State Representation}


    In the Belief-SG model, the game maintains a belief state capturing the agent’s uncertainty about the hidden aspects of the game. In GGP, this belief representation must be general, interpretable, and independent of agent-specific inference.

    Although designing a universal belief representation for all games is impractical, most games commonly played by humans share structural features that can be exploited. In particular, imperfect-information games tend to involve hidden elements in one of two forms: either unknown positions of pieces (e.g., the position of the pieces in Battleship or Kriegspiel) or unknown identities of pieces (e.g., the color and suit of a card in Poker or the rank of a piece in Stratego). In practice, the majority of games with imperfect information fall into the latter category, and this work focuses exclusively on such games.
    
    The remainder of this section describes two belief representations built on this structure. The first captures the set of logically feasible states using a constraint satisfaction formulation. The second extends this representation with probabilistic inference through Belief Propagation to estimate the likelihood of each piece's identity.

    \subsection{Constraint-Based Belief Representation}

        In imperfect-information games of the type we consider, uncertainty arises solely from the hidden identities of pieces. The rest of the game state, such as piece positions, turn order, the structure of the board, or mutable variables like scores or pots, is fully observable. We therefore divide the belief state into two components: a deterministic, observable part, and a hidden part representing the possible identities of the pieces.

        We represent the hidden components using a constraint satisfaction problem, a formalism well-suited to discrete domains. A CSP consists of a set of variables, each with a finite domain of possible values, and a set of constraints that specify which combinations of values are allowed. In this context, the CSP encodes all states of the game that remain logically feasible given public information and past actions.

        Formally, we define the hidden part of the belief state as follows:
        \begin{itemize}
            \item[--] $\mathcal{P}$ is the set of all pieces with unknown identities.
            \item[--] $\mathcal{T}$ is the set of piece types. Each type \( t \in \mathcal{T} \) is associated with a set of valid identities \( \mathcal{V}_t \).
            \item[--] $\theta \colon \mathcal{P} \to \mathcal{T}$ is the type function mapping each piece to its corresponding type.
            \item[--] $\mathcal{D} = \{ D_p \}_{p \in \mathcal{P}}$ is a set of domains, where \( D_p \subseteq \mathcal{V}_{\theta(p)} \) represents the current set of identities that piece \( p \) may assume, given the accumulated public and private information acquired in the current state.
        \end{itemize}

        To enforce consistency in the number of piece identities that can appear, we associate a Global Cardinality Constraint (GCC)~\cite{Regin_1996_Generalized} with each type. For each $t \in \mathcal{T}$, a constraint $\text{GCC}_t$ ensures that the number of occurrences of each identity $v \in \mathcal{V}_t$ among the pieces of type $t$ remains within allowed limits (e.g., the known number of cards in a deck). These constraints reflect the underlying structure of the game and are essential to rule out infeasible states.

        The overall belief representation takes the form of a collection of CSPs, one per piece type, where each CSP includes a set of variables $\mathcal{P}_t = \{ p \in \mathcal{P} \mid \theta(p) = t \}$, domains $\{ D_p \}_{p \in \mathcal{P}_t}$, and a GCC constraint $\text{GCC}_t$ over the variables.

        As the game progresses, actions can eliminate some values from the domains. For example, moving a piece in Stratego rules out the piece being a flag or a bomb. When this occurs, the GCCs propagate these changes to ensure that the overall count of identities remains feasible. This can lead to further pruning of other domains, or even force a piece to take on a specific identity when only one valid option remains.

        The resulting CSP represents the support of the belief state, that is, the set of all game states that remain logically consistent with the history of the game. Although this representation does not assign probabilities to states, it provides a compact and efficient structure to rule out impossible configurations.

    \subsection{Probabilistic Belief Representation}

        While the constraint-based belief model captures which identity assignments are still logically feasible, it provides no information about the likelihood of each possibility. In some games, this information may help agents resolve uncertainty between multiple options. To enrich the belief state with such quantitative insight, we estimate marginal probability distributions over the domain of each piece, reflecting how likely each identity is given the current state of the game.

        A naive approach would involve enumerating all valid assignments consistent with the CSP and computing exact frequencies to derive probabilities. However, this is intractable for even moderately sized games due to the exponential number of valid assignments. Instead, inspired by the work of Pesant G.~\cite{Pesant_2019_Support}, we adopt a more scalable approach based on Belief Propagation (BP), a message-passing algorithm used for approximate inference in graphical models.
        
        BP is typically applied on factor graphs~\cite{Loeliger_2004_FactorGraphs}, which are bipartite graphs composed of variable and factor nodes. Each variable node represents a discrete random variable, and factor nodes impose constraints on subsets of variables and encode their joint compatibility.
        
        BP proceeds by iteratively exchanging messages between variable and factor nodes. Each variable node sends messages to its connected factors summarizing its current belief, while each factor node responds with messages reflecting the consistency of the variable’s current belief with other connected variables under the factor's constraint. Once the messages converge, the marginal distribution of each variable can be approximated by combining its incoming messages.
        
        To apply BP to our belief state representation, we first reinterpret the CSP as a factor graph. Each piece identity variable becomes a variable node, and each constraint becomes a factor node. In particular, the GCCs introduced earlier must be translated into factor nodes. However, computing BP messages for standard multi-value GCCs directly is computationally intractable due to their complex combinatorial nature.
        
        To simplify this, we decompose each GCC into a set of simpler count constraints, one for each identity $v \in \mathcal{V}_t$. Each count constraint enforces how many times the identity $v$ can occur among variables of type $t$ and is implemented as a separate factor node connected only to the variables whose domain includes $v$.

        Each count constraint effectively acts as a sum constraint, since limiting the number of times an identity appears amounts to summing binary indicators over the variables. We adapt the message computation algorithm introduced by Pesant~\cite{Pesant_2019_Support} for such constraints, enabling efficient propagation in our model. This results in a set of approximate marginal distributions over the domains, reflecting the relative likelihood of each identity assignment given the current game state.

\section{Experiments}

    To evaluate the practical impact of probabilistic versus purely constraint-based belief representations, we compare the performance of agents using each approach across two imperfect-information games. All experiments were conducted on a cluster of Intel Xeon E5-2667 CPUs at 3.2\,GHz running Linux. The implementation is written in C++, using the Gecode library~\cite{Schulte_2010_Gecode} for constraint solving, and is publicly available\footnote{\url{https://github.com/AchilleMorenville/Belief-SG}}.

    We evaluate two general-purpose agents based on Monte Carlo simulations, both of which operate without any game-specific heuristics. Each agent is tested under two belief models: one using only constraint-based belief state representation, and the other enriched with probabilistic estimates via BP. This controlled setup isolates the effect of belief representation on decision-making performance.

    \subsection{Agents}

        Both agents handle imperfect information via determinization, sampling fully observable states consistent with the belief model. In the constraint-based setting, values are assigned uniformly at random, with constraint propagation ensuring consistency. In the probabilistic setting, sampling is guided by the marginal distributions estimated through BP. Variables are selected in order of confidence, and values are sampled from their distributions. After each assignment, the CSP is updated, and BP is incrementally rerun to maintain global consistency.

        \subsubsection{Pure Monte Carlo (PMC)}

            This agent performs a flat Monte Carlo search. For each legal action, it samples a fixed number of playouts from that action, each starting from a sampled fully observable determinization. The outcomes are averaged, and the action with the highest expected return is selected. No search tree is built, and evaluation relies entirely on rollout statistics.

        \subsubsection{Decoupled UCT (DUCT)}

            This agent builds on the Decoupled UCT algorithm for simultaneous games~\cite{Tak_2014_Monte}, an extension of the classic UCT algorithm~\cite{Kocsis_2006_Bandit}. In this formulation, each player independently selects actions at each node based on their own statistics, without modeling the opponents’ choices. To handle imperfect information, the algorithm is applied across multiple sampled determinizations, each maintaining its own search tree. The final action is selected by majority vote across all trees.

    \subsection{Games}

        \subsubsection{Mini-Stratego}

            We use a simplified version of Stratego where each player controls five hidden pieces on a $5 \times 5$ board. The possible identities are \textit{Flag}, \textit{Bomb}, \textit{Miner}, and \textit{Soldier}. The goal is to capture the flag of the opponent. Soldiers defeat Miners, and only Miners can defuse Bombs.

        \subsubsection{Goofspiel}

            Goofspiel is a simultaneous bidding game in which each player has a hand of 13 cards numbered 1 to 13, and a separate deck determines the prize cards. At each round, a prize card is revealed, and players simultaneously bid one card from their hand. The highest bidder wins the prize, with ties resulting in a shared reward. Each card may only be used once.

    \subsection{Results}

        We evaluated five agents across two imperfect-information games to compare probabilistic and constraint-based belief representations. Each agent pair played 1,000 matches, totaling 10,000 per game. Agents include PMC and DUCT variants using constraint-based (C) or probabilistic (P) beliefs, plus a random baseline (R). All used the same simulation budget: 10 determinizations per move, with PMC running 1,000 simulations per action and DUCT 1,000 per determinization. Table~\ref{tab:full-results} shows win rates for each agent against all opponents.

        \begin{table}[t]
            \centering
            \caption{Win rates (\%) of row agents against column agents.}
            \label{tab:full-results}
            \setlength{\tabcolsep}{5pt} 
            \renewcommand{\arraystretch}{1.1} 
            \begin{tabular}{llccccc}
                \toprule
                \textbf{Game} & \textbf{Agent} & $\text{R}$ & $\text{PMC}_{\text{C}}$ & $\text{PMC}_{\text{P}}$ & $\text{DUCT}_{\text{C}}$ & $\text{DUCT}_{\text{P}}$ \\
                \midrule
                \multirow{4}{*}{Mini-Stratego}
                  & $\text{PMC}_{\text{C}}$   & 86.5 & --   & 49.8 & 45.1 & 42.2 \\
                  & $\text{PMC}_{\text{P}}$   & 87.2 & 50.2 & --   & 41.4 & 45.7 \\
                  & $\text{DUCT}_{\text{C}}$  & 88.4 & 54.9 & 58.6 & --   & 49.1 \\
                  & $\text{DUCT}_{\text{P}}$  & 87.7 & 57.8 & 54.3 & 50.9 & --   \\
                \midrule
                \multirow{4}{*}{Goofspiel}
                  & $\text{PMC}_{\text{C}}$   & 91.6 & --   & 49.9 & 21.5 & 20.0 \\
                  & $\text{PMC}_{\text{P}}$   & 90.0 & 50.1 & --   & 21.3 & 19.3 \\
                  & $\text{DUCT}_{\text{C}}$  & 79.2 & 78.5 & 78.7 & --   & 49.1 \\
                  & $\text{DUCT}_{\text{P}}$  & 78.6 & 80.0 & 80.7 & 50.9 & --   \\
                \bottomrule
            \end{tabular}
        \end{table}

    \subsection{Discussion}

       The results show that constraint-based and probabilistic belief representations yield comparable performance across games and agents. In both Mini-Stratego and Goofspiel, agents using Belief Propagation show no consistent advantage over those using feasibility constraints alone. This suggests that for determinization-based agents, the added cost of probabilistic inference may be unjustified when constraint filtering already approximates the state well.

        
        Across both games, DUCT variants consistently outperform their PMC counterparts, highlighting the strength of tree-based planning under uncertainty. This finding reinforces that the planning algorithm itself has a greater influence on performance than the precision of the belief model, at least in the context of the agents and games evaluated here.
        
        The structure of the games likely influences these outcomes. In domains with prolonged uncertainty or where subtle belief differences strongly affect decisions, richer probabilistic models may yield clearer benefits. Thus, their full value may emerge only in more information-sensitive settings or with agents that use belief states more directly in planning.

\section{Conclusion}


    This work examined the practical impact of probabilistic and constraint-based belief representations in the Belief-SG model. For determinization-based agents, constraint satisfaction provides a strong approximation of hidden state uncertainty, with only marginal gains from costlier Belief Propagation. This is an early step toward assessing the practical role of belief representation. Future work will expand to more agents, games, and complex domains to better identify when probabilistic belief estimation offers an advantage.



\section*{Acknowledgment}


This article is based on the work of COST Action CA22145 - GameTable \cite{Piette_2024_GameTable}, supported by COST (European Cooperation in Science and Technology). Computational resources have been provided by the Consortium des Équipements de Calcul Intensif (CÉCI), funded by the Fonds de la Recherche Scientifique de Belgique (F.R.S.-FNRS) under Grant No. 2.5020.11 and by the Walloon Region.

\bibliographystyle{IEEEtran}
\bibliography{IEEEabrv,bibliography}

\end{document}